\documentclass[12pt, article,oneside, 
headinclude,
footinclude, 
BCOR=5mm]{scrbook}
\usepackage{setspace} 
 \makeatletter
\DeclareOldFontCommand{\rm}{\normalfont\rmfamily}{\mathrm}
 \DeclareOldFontCommand{\sf}{\normalfont\sffamily}{\mathsf}
 \DeclareOldFontCommand{\tt}{\normalfont\ttfamily}{\mathtt}
 \DeclareOldFontCommand{\bf}{\normalfont\bfseries}{\mathbf}
 \DeclareOldFontCommand{\it}{\normalfont\itshape}{\mathit}
 \DeclareOldFontCommand{\sl}{\normalfont\slshape}{\@nomath\sl}
 \DeclareOldFontCommand{\sc}{\normalfont\scshape}{\@nomath\sc}
 \makeatother
\usepackage[english]{babel}
\usepackage[T1]{fontenc} % Use 8-bit encoding that has 256 glyphs
\usepackage{times}
\usepackage[utf8]{inputenc}% Required for including letters with accents
\usepackage{newunicodechar}
\newunicodechar{〈}{\langle}
\usepackage{tabularx}
\usepackage{graphicx} % Required for including images
\graphicspath{{Figures/}} % Set the default folder for images

\usepackage{enumitem} % Required for manipulating the whitespace between and within lists
\newlist{steps}{enumerate}{1}
\setlist[steps, 1]{label = Step \arabic*:}

\usepackage{lipsum} % Used for inserting dummy 'Lorem ipsum' text into the template

\usepackage{amsmath,amssymb,amsthm} % For including math equations, theorems, %symbols, etc
\usepackage{varioref} % More descriptive referencing
\usepackage{booktabs} % For \toprule, \midrule and \bottomrule

\usepackage{siunitx} % Formats the units and values

\usepackage{pgfplotstable} % Generates table from .csv

\usepackage{etoolbox}

\usepackage{float}

\usepackage{amsmath}

\usepackage[T1]{fontenc}

\usepackage{natbib}

\usepackage{listings}

\usepackage[labelfont=bf]{caption}
\usepackage{longtable}

\usepackage{filecontents}  

\usepackage{amsfonts}

\usepackage[colorinlistoftodos]{todonotes}

\usepackage{algorithm,algpseudocode}
\usepackage{algorithmicx}

\usepackage{lineno}

\usepackage{etoolbox}

\usepackage{blindtext}

\usepackage{enumitem}
\usepackage{lmodern}
\usepackage{fullpage}
\usepackage{alltt}
\usepackage{boxedminipage}
\usepackage{color}

\usepackage{booktabs}

  \usepackage{url}
\usepackage{subfig}
\usepackage[font={normal,it}]{caption}

\theoremstyle{definition} % Define theorem styles here based on the definition style (used for definitions and examples)

\theoremstyle{plain} % Define theorem styles here based on the plain style (used for theorems, lemmas, propositions)

\theoremstyle{remark} % Define theorem styles here based on the remark style (used for remarks and notes)
\usepackage{authblk}
\usepackage{chngcntr}
\counterwithin{figure}{section}
\PassOptionsToPackage{hyphens}{url}
\usepackage{breakcites}
\usepackage{caption}
\usepackage{listings}
\usepackage{titling}
\usepackage{chngcntr}
\counterwithin{table}{section}
\date{} % An optional date to appear under the author(s)

\let\clsCenter\Center\let\clsendCenter\endCenter
\let\Center\undefined\let\endCenter\undefined
\usepackage{ragged2e}
\let\Center\clsCenter
\let\endCenter\clsendCenter

\newcommand\SentenceCase[1]{%
  \caselower{}%
  \capitalize{\thestring}%
}
\begin{singlespace}
\title{A simulated annealing algorithm for\\
joint stratification and\\
sample allocation} 
\author
{Mervyn O'Luing,$^{1}$ Steven Prestwich,$^{1}$ S. Armagan Tarim$^{2}$
}
\end{singlespace}
\usepackage{bibentry}

\pgfplotsset{compat=1.15}
\begin{document}
%\begin{titlepage}
\begin{singlespace}
\begin{minipage}[h]{\textwidth}
\textbf{\maketitle}
\end{minipage}
%----------------------------------------------------------------------------------------
%	ABSTRACT
%--------------------------------------------
\begin{center}
    %\vspace{0.9cm}
    \textbf{Abstract}
\end{center}

 % This section will not appear in the table of contents due to the star (\section*)

This study combines simulated annealing with delta evaluation to solve the joint stratification and sample allocation problem. In this problem, atomic strata are partitioned into mutually exclusive and collectively exhaustive strata. Each partition of atomic strata is a possible solution to the stratification problem, the quality of which is measured by its cost.  The Bell number of possible solutions is enormous, for even a moderate number of atomic strata, and an additional layer of complexity is added with the evaluation time of each solution. Many larger scale combinatorial optimisation problems cannot be solved to optimality, because the search for an optimum solution requires a prohibitive amount of computation time. A number of local search heuristic algorithms have been designed for this problem but these can become trapped in local minima preventing any further improvements. We add, to the existing suite of local search algorithms, a simulated annealing algorithm that allows for an escape from local minima and uses delta evaluation to exploit the similarity between consecutive solutions, and thereby reduces the evaluation time. We compared the simulated annealing algorithm with two recent algorithms. In both cases, the simulated annealing algorithm attained a solution of comparable quality in considerably less computation time.  
\\
\\
\textbf{Keywords:} Simulated annealing algorithm; Optimal stratification; Sample
allocation; R software.

{\let\thefootnote\relax\footnotetext{\textsuperscript{1} \textit{Insight Centre for Data Analytics, Department of Computer Science, University College Cork, Ireland.} Email: {mervyn.oluing@insight-centre.org},{ steven.prestwich@insight-centre.org }}}

{\let\thefootnote\relax\footnotetext{\textsuperscript{2} \textit{Cork University Business School, University College Cork, Ireland.} Email: {armagan.tarim@ucc.ie}}}

\end{singlespace}
%\end{titlepage}
%----------------------------------------------------------------------------------------

%----------------------------------------------------------------------------------------
%	INTRODUCTION
%----------------------------------------------------------------------------------------

\section{Introduction}

In stratified simple random sampling, a population is partitioned into mutually exclusive and collectively exhaustive strata, and then sampling units from each of those strata are randomly selected. The purposes for stratification are discussed in \cite{william1977sampling}. If the intra-strata variances were minimized then precision would be improved. It follows that the resulting small samples from each stratum can be combined to give a small sample size. \par To this end, we intend to construct strata which are internally homogeneous but which also accommodate outlying measurements. To do so, we adopt an approach which entails searching for the optimum partitioning of \emph{atomic strata} (however, the methodology can also be applied to \emph{continuous} strata) created from the Cartesian product of categorical stratification variables, see \cite{benedetti2008tree,ballin2013joint,ballin2020optimization}.\par The Bell number, representing the number of possible partitions (stratifications) of a set of atomic strata, grows very rapidly with the number of atomic strata \citep{ballin2013joint}.  In fact, there comes a point where, even for a moderate number of atomic strata and the most powerful computers, the problem is intractable, i.e. there are no known efficient algorithms to solve the problem. \par Many large scale combinatorial optimisation problems of this type cannot be solved to optimality, because the search for an optimum solution requires a prohibitive amount of computation time. This compels one to use \emph{approximisation algorithms} or \emph{heuristics} which do not guarantee optimal solutions, but can provide approximate solutions in an acceptable time interval. In this way, one trades off the quality of the final solution against computation time \citep{van1987simulated}. In other words, heuristic algorithms are developed to find a solution that is "good enough" in a computing time that is "small enough" \citep{sorensen2013metaheuristics}. \par A number of heuristic algorithms have been developed to search for optimal or near optimal solutions, for both univariate and multivariate scenarios of this problem. This includes the hierarchichal algorithm proposed by \cite{benedetti2008tree}, the genetic algorithm proposed by \cite{ballin2013joint} and the grouping genetic algorithm proposed by \cite{oluing2019grouping}. Although effective, the evaluation function in these algorithms can be costly in terms of running time. \par We add to this work with a simulated annealing algorithm (SAA) \citep{kirkpatrick1983optimization,vcerny1985thermodynamical}. SAAs have been found to work well in problems such as this, where there are many local minima and finding an approximate global solution in a fixed amount of computation time is more desirable than finding a precise local minimum \citep{takeang2019multiple}. We present a SAA to which we have added delta evaluation (see section \ref{improving}) to take advantage of the similarity between consecutive solutions and help speed up computation times. \par We compared the performance of the SAA on atomic strata with that of the grouping genetic algorithm (GGA) in the \emph{SamplingStrata} package \citep{ballin2020optimization}. This algorithm implements the grouping operators described by \cite{oluing2019grouping}. To do this, we used sampling frames of varying sizes containing what we assume to be completely representative details for target and auxiliary variable columns. \par Further to the suggestion of a \emph{Survey Methodology} reviewer, we subsequently compared the SAA with a traditional genetic algorithm (TGA) used by \cite{ballin2020optimization} on continuous strata. In both sets of experiments, we used an initial solution created by the k-means algorithm \citep{hartigan1979algorithm} in a two-stage process (see section \ref{two-stage_} for more details). \par Section \ref{background} provides background information on atomic strata, introduces the SAA and motivates the addition of delta evaluation as a means to improve computation time. Two-stage simulated annealing is also discussed.  Section \ref{joint} of the paper describes the cost function and evaluation algorithm. Section \ref{outline} provides an outline of the SAA. Section \ref{improving} presents the improved SAA with delta evaluation.  Section \ref{experiments} provides a comparison of the performance of the SAA with the GGA using an initial solution and fine-tuned hyperparameters. Section \ref{comparison_cont} then provides details of the comparison of the SAA with the genetic algorithm in \cite{ballin2020optimization} on continuous strata. Section \ref{conclusions} presents the conclusions and section \ref{further} suggests some further work. The appendix - section \ref{Appendix} contains background details on precision constraints, the hyperparameters, and the process of fine-tuning the hyperparameters for both comparisons as well as the computer specifications. 

\section{Background information} \label{background}

\subsection{Stratification of atomic strata}\label{Strat_atom}

Atomic strata are created using categorical auxiliary variable columns such as \emph{age group}, \emph{gender} or \emph{ethnicity} for a survey of people or \emph{industry}, \emph{type of business} and \emph{employee size} for business surveys. The cross-classification of the class-intervals of the auxiliary variable columns form the atomic strata. \par
Auxiliary variable columns which are correlated to the target variable columns may provide a gain in sample precision or \emph{similarity}.  Each target variable column, $y_g$, contains the value of the survey characteristic of interest, e.g. \emph{total income}, for each population element in the sample. \par 
Once these are created, we obtain summary statistics, such as the number, mean and standard deviation of the relevant observed values, from the one or more target variable columns that fall within each atomic stratum. The summary information is then aggregated in order to calculate the means and variances for each stratum which in turn are used to calculate the sample allocation for a given stratification.   \par 
The partitioning of atomic strata that provides the \emph{global minimum} sample allocation, i.e. the minimum of all possible sample allocations for the set of possible stratifications, is known as an \emph{optimal stratification}. There could be a multiple of such partitionings. Although an optimum stratification is \emph{the} solution to the problem, each stratification represents a solution of varying quality (the lower the cost (\emph{minimum} or \emph{optimal} sample allocation) the higher the quality). For each stratification, the cost is estimated by the Bethel-Chromy algorithm \citep{bethel1985optimum,bethel1989sample,chromy1987design}. A more detailed description, and discussion of the methodology for this approach for joint determination of stratification and sample allocation, can be found in \cite{ballin2013joint}.  

\subsection{Simulated annealing algorithms} \label{SA}

The basic principle of the SAA \citep{kirkpatrick1983optimization,vcerny1985thermodynamical} is that it can accept solutions that are inferior to the current best solution in order to find the global minima (or maxima). It is one of several stochastic local search algorithms, which focus their attention within a local neighbourhood of a given initial solution \citep{cortez2014modern}, and use different stochastic techniques to escape from attractive local minima \citep{hoos2004stochastic}. \par
Based on physical annealing in metallurgy, the SAA is designed to simulate the controlled cooling process from liquid metal to a solid state \citep{Luke2013Metaheuristics}. This controlled cooling uses the temperature parameter to compute the probability of accepting inferior solutions \citep{cortez2014modern}. This acceptance probability is not only a function of the temperature, but also the difference in cost between the new solution and the current best solution. For the same difference in cost, a higher temperature means a higher probability of accepting inferior solutions. 
\par For a given temperature, solutions are iteratively generated by applying a small, randomly generated, perturbation to the current best solution. Generally, in SAAs, a perturbation is the small displacement of a randomly chosen particle \citep{van1987simulated}. In the context of our problem, we take perturbation to mean the displacement (or re-positioning) of $q$ (generally $q=1$) randomly chosen atomic strata from one randomly chosen stratum to another. \par With a perturbation, the current best solution transitions to a new solution. If a perturbation results in a lower cost for the new solution, or if there is no change in cost, then that solution is always selected as the current best solution. If the new solution results in a higher cost, then it is accepted at the above mentioned acceptance probability. This acceptance condition is called the Metropolis criterion \citep{metropolis1953equation}. This process continues until the end of the sequence, at which point the temperature is decremented and a new sequence begins. \par If the perturbations are minor, then the current solution and the new solution will be very similar. Indeed, in our SAA we are assuming only a slight difference between consecutive solutions owing to such perturbations (see section \ref{Perturbation} for more details).  For this reason we have added delta evaluation, which will be discussed further in section \ref{improving}, to take advantage of this similarity and help improve computation times. \par Accordingly, and as mentioned in the introduction, we present a SAA with delta evaluation and compare it with the GGA when both are combined with an initial solution. We also compare it with a genetic algorithm used by \cite{ballin2020optimization} on continuous strata. We provide more background details on initial solutions in section \ref{two-stage_} below.

\subsection{Two-stage simulated annealing}\label{two-stage_}

A two stage simulated annealing process, where an initial solution is generated by a heuristic algorithm in the first stage, has been proposed for problems such as the \emph{cell placement problem} \citep{grover1987standard,rose1988parallel} or the \emph{graph partitioning problem} \citep{johnson1989optimization}. \cite{lisic2018optimal} combined an initial solution, generated by the k-means algorithm, with a simulated annealing algorithm, for a problem similar in nature to this problem, but where the sample allocation as well as strata number are fixed, and the algorithm searches for the optimal arrangement of sampling units between strata.\par
The simulated annealing algorithm used by \cite{lisic2018optimal} starts with an initial solution (stratification and sample allocation to each stratum) and, for each iteration, generates a new candidate solution by moving one atomic stratum from one stratum to another and adjusting the sample allocation for that stratification. Each candidate solution is then evaluated to measure the coefficient of variation (CV) of the target variables and is accepted, as the new current best solution, if its objective function is less than the preceding solution. Inferior quality solutions are also accepted at a probability, $\rho$, which is a function of a tunable temperature parameter and the change in solution quality between iterations. The temperature cools, at a rate which is also tunable, as the number of iterations increases.\par
Following this work, \cite{ballin2020optimization} recommended combining an initial solution, generated by k-means, with the grouping and traditional genetic algorithms. They demonstrate that the k-means algorithm provides better starting solutions when compared with the starting solution generated by a stochastic approach. We also combine a k-means initial solution with the SAA in the experiments described in sections \ref{experiments} and \ref{comparison_cont}. 

\section{The joint stratification and sample allocation problem} \label{joint}
 
Our aim is to partition $L$ atomic strata into $H$ non-empty sub-populations or strata. A partitioning represents a stratification of the population. We aim to minimise the sample allocation to this stratification while keeping the measure of similarity less than or equal to the upper limit of precision, $\varepsilon_g$. This similarity is measured by the CV of the estimated population total for each one of $G$ target variable columns, $\hat{T}_g$. We indicate by $n_h$ the sample allocated to stratum $h$ and the survey cost for a given stratification is calculated as follows:

\begin{equation}
C\left(n_1,\ldots,n_H\right)= \sum_{h=1}^{H}C_hn_h\notag
\end{equation}
where $C_h$ is the average cost of surveying one unit in stratum $h$ and $n_h$ is the sample allocation to stratum $h$.  In our analysis $C_h$ is set to $1$. \par
The variance of the estimator is given by:% \citep{oluing2019grouping}:
\begin{equation*}
\begin{array}{rrclcl}
\mbox{VAR}\left(\hat{T}_g\right)=\sum_{h=1}^{H}N_h^2\left(1-\frac{n_h}{N_h}\right) 
\frac{S_{h,g}^2}{n_h} \;\;\; (g=1,\ldots,G)\label{variance_1}
\end{array}
\end{equation*}

where $N_h$ is the number of units in stratum $h$ and $S_{h,g}^2$ is the variance of stratum $h$ for each target variable column $g$.

As mentioned above $\varepsilon_g$ is the upper precision limit for the CV for each $\hat{T}_g$:
\begin{equation*}
CV( \hat{T}_{g}) =    \frac{\sqrt{VAR (\hat{T}_{g})}} {E(\hat{T}_{g})}  \leq \varepsilon_g\;.         
\end{equation*}
The problem can be summarised in this way:
\begin{equation*}
\begin{array}{l}
\min \quad \quad \quad \quad n = \sum_{h=1}^{H}n_h \\
\text{subject to } \quad CV\left( \hat{T}_g\right)\leq \varepsilon_g\notag \;\;\; (g=1,\ldots,G)\;.
\end{array}
\end{equation*}
To solve the allocation problem for a particular stratification with the Bethel-Chromy algorithm the upper precision constraint for variable $g$ can be expressed as follows:

\begin{equation*}
\begin{array}{rrclcl}
CV\left(\hat{T}_g\right)^2 \leq \varepsilon_g^2\notag
\equiv \sum_{h=1}^{H}\frac{N_h^2 S_{h,g}^2}{n_h} -N_h S_{h,g}^2 \leq E(\hat{T}_g^2) \varepsilon_g^2
\end{array}
\end{equation*}

\begin{equation*}
\begin{array}{rrclcl}
					\equiv \sum_{h=1}^{H}\frac{N_h^2 S_{h,g}^2}{\left(E(\hat{T}_g^2) \varepsilon_g^2 + \sum_{h=1}^{H} N_h S_{h,g}^2\right)\ n_h}  \leq 1.
\end{array}
\end{equation*}

Then we substitute $\frac{N_h^2 S_{h,g}^2}{	\left( E(\hat{T}_g^2)\varepsilon_g^2 + \sum_{h=1}^{H} N_h S_{h,g}^2\right)}$ with $\xi{_h,g}$ and replace the problem summary with the following:

\begin{equation*}
\begin{array}{l}
\min n = \sum_{h=1}^{H}n_h\\
\sum_{h=1}^{H}\frac{\xi{_h,g}}{n_h}\  \leq 1 \;\;\; (g=1,\ldots,G)\;.
\end{array}
\end{equation*}

where $\frac{1}{n_h}>0$. The Bethel-Chromy algorithm uses Lagrangian multipliers to derive a solution for each $n_h$. 

\begin{equation*}
\frac{1}{n_h}=
\begin{cases}
  \frac{\sqrt{1}}{(\sqrt{\sum_{g=1}^{G}\alpha_g \xi_h,g}\sum_{h=1}^{H}\sqrt{\sum_{g=1}^{G}\alpha_g \xi{_h,g}})\ }\text{ if $\sum_{g=1}^{G}\alpha_g \xi_h,g > 0 $}\\      
  + \infty  			 \text{ otherwise}\;.\\
\end{cases}
\end{equation*}

where $\alpha_g = \frac{\lambda _g}{\sum_{g=1}^{G} \lambda_g}$, and $\lambda _g$ is the Lagrangian multiplier \citep{benedetti2008tree}. The algorithm starts with a default setting for each $\alpha_g$ and uses gradient descent to converge to a final value for them. 

\section{Outline of the simulated annealing algorithm}\label{outline} 
The SAA with delta evaluation is described in algorithm \ref{SAA} below. We then describe the heuristics we have used in the SAA. Delta evaluation is explained in more detail in section \ref{improving}.

\begin{singlespace}
\begin{algorithm}[H]
\caption{Simulated annealing algorithm}
\label{SAA}
\scriptsize
\begin{algorithmic}
\begin{singlespace}
\Function{SimulatedAnnealing}{$S$ is the starting solution, $f$ is the evaluation function (Bethel-Chromy algorithm), $best$ is the current best solution, $BSFSF$ is the best solution found so far,  $maxit$ is the maximum number of sequences, $J$ is the length of sequence, $T_{max}$ is the starting temperature, $T_{min}$ is the minimum temperature, $DC$ is the Decrement Constant, $L_{max \%}$ is a \% of $L$ (number of atomic strata), $P(H+1)$ is the probability of a new stratum, $H+1$, being added}
\State $T\gets T_{max}$
\State $best\gets$ S
\State $Cost(best)\gets f(best)$ \Comment{using Bethel-Chromy algorithm}
\While{$ i<maxit$ \&\& $T>T_{min}$}
\If{\Call{random}{$0,1$} $\leq 1/J$} 
\For{$ l = 1$ to $L$}
\If{\Call{random}{$0,1$} $\leq P(H+1)$}\\\quad \quad \quad \quad \quad \quad \quad move atomic stratum  $l$ to new stratum $H+1$ \Comment{see section \ref{sequences} }
\EndIf
\EndFor
\EndIf
\For{$ j = 1$ to $J$}
    \If{$i = 1$ \& $j =1$} $q = L \times L_{max \%}$
    \ElsIf{$i=1$ \& $j > 1$} $q =ceiling(q \times 0.99)$ \Comment{0.99 is not tunable}
    \ElsIf{$i > 1$} $q = 1$ 
    \EndIf
    \State Randomly select $h$ and $h'$
	\State $next\gets $ \textbf{PERTURBATION}($best$) \\ \Comment{Assign $q$ atomic strata from $h$ to $h'$}
	\State $Cost(next)\gets f(next)$ \Comment{using delta evaluation}
	\State $\Delta E \gets $\textbf{COST}($next$) $-$ \textbf{COST}($best$)
	\If{$\Delta E \leq 0$}
      \State $best\gets next$ 
	\ElsIf{\Call{random}{$0,1$} $<$ $e^{(-\frac{\Delta E}{T})}$} \Comment{Metropolis Criterion}
		\State $best\gets next$
	\EndIf
	\If{best $<=$ $BSFSF$}
	       \State $ BSFSF \gets best$ 
	 \EndIf
\EndFor
\State  	$T \gets T * DC$\\
\EndWhile  
\State \Return \textsc{$BSFSF$}
\EndFunction
\end{singlespace}
\end{algorithmic}
\end{algorithm}
\end{singlespace}

\subsection{Perturbation}\label{Perturbation}

Consider the following solution represented by the stratification:

\begin{center}
 $\{1,3\}$, $\{2\}$, $\{4,5,6\}$
 \end{center}

The integers within each stratum represent atomic strata. In perturbation, the new solution below is created by arbitrarily moving atomic strata, in this example $q=1$, from one randomly chosen stratum to another. 
 
\begin{center}
$\{1,3,2\}$ , $\{\emptyset\}$, $\{4,5,6\}$\\
\end{center}
 
The first stratum gains an additional atomic stratum $\{2\}$ to become $\{1,3,2\}$, whereas the middle or second stratum has been "emptied" (and is deleted), and there remains only two strata. Strata are only emptied when the last remaining atomic stratum has been moved to another stratum.\par To clarify how this works in the algorithm: each solution is represented by a vector of integers - atomic strata which have the same integer are in the same stratum. A separate vector of the unique integers in the solution represents the strata.  For example, the first solution  $\{1,3\}$, $\{2\}$, $\{4,5,6\}$ would be represented by the vector $[1 \quad 2\quad 1\quad3 \quad3 \quad3]$ and the strata would be represented by the vector $[1 \quad 2\quad 3]$. When the new solution is created, the second stratum has been removed and is no longer part of the solution. That is to say, the vector for the new solution is: $[1 \quad 1\quad 1\quad3 \quad3 \quad3]$ and the strata vector is $[1 \quad 3]$. With stratum 2 removed, and for clarity, we rename stratum $3$ to $2$ so that this solution becomes:  $[1 \quad 1\quad 1\quad2 \quad2 \quad2]$, and the strata are now represented by the vector  $[1 \quad 2]$. Strata $[1 \quad 2]$ will remain in any further solutions unless another stratum is "emptied" or a new stratum is added. 

\subsection{Evaluation and acceptance}

Each new solution is evaluated using the Bethel-Chromy algorithm and the Metropolis acceptance criterion is applied. If accepted, the new solution differs from the previous solution only by the above mentioned perturbation. If it is not accepted, we continue with the previous solution, and again try moving $q$ randomly chosen atomic strata between two randomly selected strata. 

\subsection{Sequences and new strata}\label{sequences} This continues for the tunable length of the sequence, $J$. This should be long enough to allow the sequence to reach equilibrium. However, there is no rule to determine $J$. At the commencement of each new sequence, we have $H$ strata in the current best solution. With a fixed probability of $1/J$, an additional stratum is added.  If a new stratum is to be added, the SAA loops through each atomic stratum and moves it to a new stratum, which is called $H+1$, because each stratum is labelled sequentially from $1$ to $H$ (see section \ref{Perturbation}), at a tunable probability, $P(H+1)$. The algorithm runs for a tunable number of sequences, $maxit$.
\subsection{Temperature}
The temperature is decremented from a starting temperature, $T_{max}$, to a minimum temperature, $T_{min}$, or until $maxit$ has been reached. As we are starting with a near optimal solution, we select $T_{max}$ as no greater than $0.01$ and we set $T_{min}$ to be $1.0 \times 10^{-11}$. \par This is to allow for the advanced nature of the search, and allows the algorithm to focus more on the search for superior solutions, with an ever-reducing probability of accepting inferior solutions. However, a low temperature, $T$, does not always equate to a low probability of acceptance. \par
\begin{figure}[H]
\centering 
\includegraphics[]{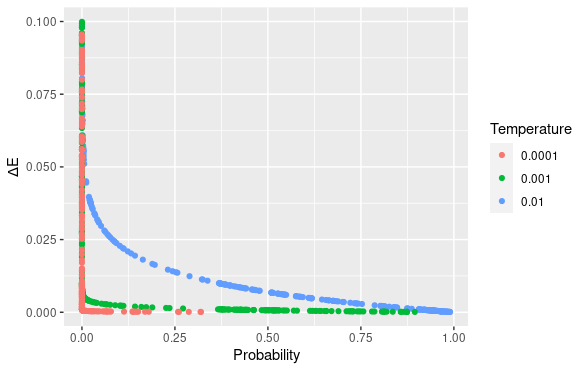}
\caption{\textbf{Probability of accepting an inferior solution as a function of $\Delta E $ and $T$}}
\label{probability}
\end{figure}

 Small positive differences in solution quality (where the new solution has a marginally inferior quality to the current best solution), $\Delta E $, occur often because we are starting with a good quality initial solution. Figure \ref{probability} demonstrates the probability of such solutions being accepted, $e^{(-\frac{\Delta E}{T})}$, increases the smaller this difference becomes for the same $T$. Nonetheless, figure \ref{probability} also demonstrates that for the same changes in solution quality as the $T$ decreases, the probability also decreases  (and it behaves increasingly like a hill climbing algorithm). 

\section{Improving the performance of the simulated annealing algorithm using delta evaluation} \label{improving}
As outlined earlier, the only difference between consecutive solutions is that $q$ atomic strata have been moved from one group into another. As with the other heuristics, $q$ is also tunable, and for the first sequence we have added the option of setting $q > 1$ and reducing $q$ for each new solution in the first sequence until $q = 1$. The reason for this is that, where $q > 1$, the increased size of the perturbation can help reduce the number of strata. In this case, we set $q$ as a tunable percentage of the solution size, or of the number of atomic strata, $L$, to be partitioned. After the first sequence $q=1$. \par Furthermore, as the strata are mutually exclusive, this movement of $q$ atomic strata from one stratum to another does not affect the remaining strata in any way. \cite{ross1994improving} introduce a technique called delta evaluation, where the evaluation of a new solution makes use of previously evaluated similar solutions, to significantly speed up evolutionary algorithms/timetabling experiments. We use the similar properties of two consecutive solutions to apply delta evaluation to the SAA. It follows, therefore, that in the first sequence $q$ should be kept low and the reduction to $q=1$ should be swift. \par The Bethel-Chromy algorithm requires the means and variances for each stratum in order to calculate the sample allocation. However, we use the information already calculated for the remaining $H-2$ strata, and simply calculate for the two strata affected by the perturbation. Thus, the computation for the means and variances of the $H$ strata is reduced to a mere subset of that otherwise required. \par Now recall that the Bethel-Chromy algorithm starts with a default value for each $\alpha_g$, and uses gradient descent to find a final value for each $\alpha_g$. This search continues up to when the algorithm reaches a minimum step-size threshold, or alternatively exceeds a maximum number of iterations. This minimum threshold is characterised by $\epsilon$, which is set as $1.0 \times 10^{-11}$ in \cite{ballin2020optimization}, and the maximum number of iterations is $200$. We make the assumption that this search will be substantially reduced if we use the $\alpha_g$ values from the evaluation of the current solution as a starting point for the next solution. \par The above two implementations of delta evaluation result in a noticeable reduction in computation times as demonstrated in the experiments described below.

\section{Comparing the performance of the two algorithms} \label{experiments}

\subsection{Evaluation plan}\label{evalplan}

In this section, we outline the comparison of the performance of the grouping genetic algorithm with the simulated annealing algorithm. We used a number of data sets of varying sizes in these experiments. There are a number of regions in each data set (labelled here as domains). An optimal stratification and minimum sample allocation was selected for each domain. 
\par The sum of the samples for all domains provides the total sample size. The sample size, or cost of the solution, defines the solution quality. For more details on domains refer to \cite{ballin2013joint}. The aim of these experiments was to consider whether the SAA can attain comparable solution quality with the GGA in less computation time per solution thus resulting in savings in execution times. \par However, we also compared the total execution times as this is a consequence of the need to train the hyperparameters for both algorithms. More details are available in the appendix - see section \ref{finetune}. \par We tabulate the results of these experiments in section \ref{results_section} where for comparison purposes we express the SAA results as a ratio of those for the GGA. \par

\subsection{Comparing the number of solutions generated}
After the first iteration the GAA retains the elite solutions, $E$, from the previous iteration. These are calculated by the product of the elitism rate (the proportion of the chromosome population which are elite solutions), $E_{R}$, and the chromosome population size (the number of candidate solutions in each iteration), $N_{P}$. As $E$ have already been evaluated they are not evaluated again. \par For this reason, we compared the evaluation times for the evaluated solutions in the GGA with all those of the SAA. For the GGA, the total number of evaluated solutions, $N_{GGA sol}$, is a function of the number of domains, $D$, the chromosome population size, the non-elite solutions (calculated by the product of $1-E_{R}$ and $N_{P}$), and the number of iterations, $I$. For more details on the implementation of GGAs (e.g. elite solutions, elitism rate, chromosome population) we refer the reader to \citep{falkenauer1998genetic}.

\begin{equation*}
\centering
N_{GGA sol} =  (D  \times  (N_{P}+(N_{P} \times (1-E_{R})  \times (I-1))))
\end{equation*}
For the simulated annealing algorithm, the maximum number of solutions, $N_{SAA sol}$, is the number of domains, $D$, by the number of sequences, $maxit$, by the length of sequence, $J$. Recall that the SAA also stops if the minimum temperature has been reached - hence we refer to the \emph{maximum} number of solutions rather than the \emph{total}. For comparability purposes however, because the temperature is decremented only at the end of each sequence and we have a small number of sequences in the experiments below we assume the full number of solutions has been generated. 
\begin{equation*}
\centering
N_{SAA sol} = D  \cdot  maxit  \cdot  J
\end{equation*}

\subsection{Data sets, target and auxiliary variables}

Table \ref{Descriptions} provides a summary by data set of the target and auxiliary variables. 
\begin{table}[H]
\tiny
\caption{\textbf{Summary by data set of the target and auxiliary variables}}
\label{Descriptions}
\begin{tabular}{|l|l|l|l|l|}
\hline
\textbf{Dataset} & \textbf{Target variables} & \textbf{Description} & \textbf{Auxiliary variables} & \textbf{Description} \\ \hline
\textbf{SwissMunicipalities} & Surfacebois & wood area & POPTOT & total population \\ \hline
\textbf{} & Airbat & area with buildings & Hapoly & municipality area \\ \hline
\textbf{\begin{tabular}[c]{@{}l@{}}American Community \\ Survey, 2015\end{tabular}} & HINCP & \begin{tabular}[c]{@{}l@{}}Household income \\ past 12 months\end{tabular} & BLD & Units in structure \\ \hline
\textbf{} & VALP & Property value & TEN & Tenure \\ \hline
\textbf{} & SMOCP & \begin{tabular}[c]{@{}l@{}}Selected monthly owner \\ costs\end{tabular} & WKEXREL & \begin{tabular}[c]{@{}l@{}}Work experience of \\ householder and spouse\end{tabular} \\ \hline
\textbf{} & INSP & \begin{tabular}[c]{@{}l@{}}Fire/hazard/flood \\ insurance yearly amount\end{tabular} & WORKSTAT & \begin{tabular}[c]{@{}l@{}}Work status of householder \\ or spouse in family households\end{tabular} \\ \hline
\textbf{} &  &  & HFL & House heating fuel \\ \hline
\textbf{} &  &  & YBL & When structure first built \\ \hline
\textbf{US Census, 2000} & HHINCOME & total household income & PROPINSR & annual property insurance cost \\ \hline
\textbf{} &  &  & COSTFUEL & annual home heating fuel cost \\ \hline
 &  &  & COSTELEC & annual electricity cost \\ \hline
 &  &  & VALUEH & House value \\ \hline
\textbf{Kiva Loans} & term\_in\_months & \begin{tabular}[c]{@{}l@{}}duration for which \\ the loan was disbursed\end{tabular} & sector & high level categories, e.g. food \\ \hline
\textbf{} & lender\_count & \begin{tabular}[c]{@{}l@{}}the total number \\ of lenders\end{tabular} & currency & currency of the loan \\ \hline
 & loan & the amount in USD & activity & \begin{tabular}[c]{@{}l@{}}more granular category, \\ e.g.  fruits \& vegetables\end{tabular} \\ \hline
 &  &  & region & region name within the country \\ \hline
 &  &  & partner\_id & ID of the partner organization \\ \hline
\textbf{\begin{tabular}[c]{@{}l@{}}UN Commodity Trade \\ Statistics data\end{tabular}} & trade\_usd & \begin{tabular}[c]{@{}l@{}}value of the trade \\ in USD\end{tabular} & commodity & \begin{tabular}[c]{@{}l@{}}type of commodity \\ e.g. "Horses, live except \\ pure-bred breeding"\end{tabular} \\ \hline
 &  &  & flow & \begin{tabular}[c]{@{}l@{}}whether the commodity \\ was an import, export, \\ re-import or re-export\end{tabular} \\ \hline
 &  &  & category & \begin{tabular}[c]{@{}l@{}}category of commodity, \\ e.g. silk or fertilisers\end{tabular} \\ \hline
\end{tabular}
\end{table}

The target and auxiliary variables for the Swiss Municipalities data set were selected based on the experiment described in \cite{ballin2020optimization}. Accordingly, \emph{POPTOT} and \emph{HApoly} were converted into categorical variables using the k-means clustering algorithm. However, we used more domains and iterations in our experiment. More information on this data set is provided by \cite{SamplingStrata}. \par
For the remaining experiments we selected target and auxiliary variables which we deemed likely to be of interest to survey designers. Further details on the American Community Survey, 2015 \citep{USCB:2016a}, the US Census, 2000 \citep{ruggles2015integrated}, Kiva Loans \citep{kiva_2018}, and the UN commodity trade statistics data \citep{nations_2017} metadata are available in \cite{oluing2019grouping}.\par 
A further summary by data set of the number of records and atomic strata, along with a description of the domain variable, is provided in table \ref{variables_b} below. 

\begin{table}[H]
\caption{\textbf{Summary by data set of the number of records and atomic strata and a description of the domain variable}}
\label{variables_b}
\tiny
\centering
\begin{tabular}{|l|l|l|l|}
\hline
Data set                                      & \textbf{\begin{tabular}[c]{@{}l@{}}Number of  \\ records\end{tabular}} & \textbf{\begin{tabular}[c]{@{}l@{}}Number of \\ atomic strata, L\end{tabular}} & \textbf{\begin{tabular}[c]{@{}l@{}}Domain \\ variable\end{tabular}} \\ \hline
\textbf{Swiss Municipalities}                & 2,896                                                                  & 579                                                                            & REG                                                                                                                      \\ \hline
\textbf{American Community  Survey, 2015}    & 619,747                                                                & 123,007                                                                        & ST (the 51 states)                                                                                                  \\ \hline
\textbf{US Census, 2000}                     & 627,611                                                                & 517,632                                                                        & REGION                                                                                                             \\ \hline
\textbf{Kiva Loans}                          & 614,361                                                                & 84,897                                                                         & country code                                                                                                         \\ \hline
\textbf{UN Commodity Trade  Statistics data} & 352,078                                                                & 351,916                                                                        & country or area                                                                                                      \\ \hline
\end{tabular}

\end{table}

\subsection{Results}\label{results_section}

As mentioned previously, we used an initial solution in each experiment that is created by the \emph{KmeansSolution} algorithm \citep{ballin2020optimization}. We then compared the performance of the algorithms in terms of average computation time (in seconds) per solution and solution quality. Table \ref{results} provides the sample size, execution times and total execution times for the SAA and GGA. 

\begin{table}[H]
\caption{\textbf{Summary by data set of the sample size and evaluation time for the grouping genetic algorithm and simulated annealing algorithm}}
\label{results}
\centering
\tiny
\begin{tabular}{|l|l|l|l|l|l|l|}
\hline
\textbf{}                                                                           & \multicolumn{3}{l|}{\textbf{GGA}} & \multicolumn{3}{l|}{\textbf{SAA}} \\ \hline
\textbf{Data set} &
  \textbf{\begin{tabular}[c]{@{}l@{}l@{}}Sample\\ size\end{tabular}} &
  \textbf{\begin{tabular}[c]{@{}l@{}l@{}}Execution \\ time\\(seconds)\end{tabular}} &
  \textbf{\begin{tabular}[c]{@{}l@{}l@{}l@{}}Total\\ Execution \\ time\\(seconds)\end{tabular}} &
  \textbf{\begin{tabular}[c]{@{}l@{}l@{}}Sample \\ size\end{tabular}} &
  \textbf{\begin{tabular}[c]{@{}l@{}l@{}}Execution \\ time\\(seconds)\end{tabular}} &
  \textbf{\begin{tabular}[c]{@{}l@{}l@{}l@{}}Total\\ Execution \\ time\\(seconds) \end{tabular}} \\ \hline
\textbf{Swiss Municipalities}               & 128.69    & 753.82       & 10,434.30  & 125.17    & 248.91       & 8,808.63  \\ \hline
\textbf{\begin{tabular}[c]{@{}l@{}l@{}}American Community\\ Survey, 2015\end{tabular}}         & 10,136.50 & 13,146.25    & 182,152.46 & 10,279.44 & 517.76       & 6,822.42  \\ \hline
\textbf{US Census, 2000}                          & 228.81    & 2,367.36     & 36,298.35  & 224.75    & 741.75       & 8,996.85  \\ \hline
\textbf{Kiva Loans}                         & 6,756.19  & 15,669.11    & 288,946.79 & 6,646.67  & 664.30       & 7,549.87  \\ \hline
\textbf{\begin{tabular}[c]{@{}l@{}l@{}}UN Commodity Trade \\ Statistics data\end{tabular}} & 3,216.68  & 6,535.97     & 88,459.22  & 3,120.07  & 1,169.26     & 12,161.80 \\ \hline
\end{tabular}
\end{table}
The total execution time is the sum of the execution times for 20 evaluations of the GGA and SAA algorithms (by the \emph{MBO} (model-based optimisation) function in the R package \emph{mlrMBO} \citep{bischla2017mlrmbo}) using 20 sets of selected hyperparameters (i.e. one set for each evaluation). Details on the precision constraints and hyperparameters for each experiment can be found in the appendix - section \ref{Appendix}. Table \ref{results_ratio} expresses the SAA results as a ratio of those for the GGA. 
\begin{table}[H]
\caption{\textbf{Ratio comparison of the sample sizes, execution times, and total execution times for the grouping genetic algorithm and simulated annealing algorithm}}
\label{results_ratio}
\centering
\tiny
\begin{tabular}{|l|l|l|l|}
\hline
\textbf{Data set} &
  \textbf{\begin{tabular}[c]{@{}l@{}l@{}}Sample \\ size\end{tabular}} &
  \textbf{\begin{tabular}[c]{@{}l@{}l@{}}Execution \\ time\\ (seconds)\end{tabular}} &
  \textbf{\begin{tabular}[c]{@{}l@{}l@{}l@{}l@{}}Total \\execution \\ time \\(seconds)\end{tabular}} \\ \hline
\textbf{Swiss Municipalities}               & .97  & .33 & .84 \\ \hline
\textbf{\begin{tabular}[c]{@{}l@{}l@{}l@{}}American Community\\ Survey, 2015\end{tabular}}         & 1.01 & .04 & .04 \\ \hline
\textbf{US Census, 2000}                          & .98  & .31 & .25 \\ \hline
\textbf{Kiva Loans}                         & .98  & .04 & .03 \\ \hline
\textbf{\begin{tabular}[c]{@{}l@{}l@{}}UN Commodity Trade \\ Statistics data\end{tabular}} & .97  & .18 & .14 \\ \hline
\end{tabular}
\end{table}
As can be seen, the sample sizes are similar, however, the SAA shows significantly lower execution and total execution times. When these experiments are run in parallel, for cases where there is a large number of domains, there may not be enough cores to cover all domains in one run. Indeed, it may take several parallel runs to complete the task, and this will affect mean evaluation time. The computer specifications are provided in table \ref{specifications}.  Table \ref{avtime} shows the number of solutions evaluated by each algorithm to obtain the results shown in table \ref{results}. It also provides a ratio comparison of the average execution time (in seconds) per solution. 

\begin{table}[H]
\caption{\textbf{Number of solutions and ratio comparison of execution time (per second) between the grouping genetic algorithm and simulated annealing algorithm}}
\label{avtime}
\centering
\tiny
\begin{tabular}{llll}
\cline{1-3}
\multicolumn{1}{|l|}{}                                            & \multicolumn{2}{l|}{\textbf{Number of solutions evaluated}}           &                                          \\ \cline{1-3}
\multicolumn{1}{|l|}{\textbf{Data set}}                            & \multicolumn{1}{l|}{\textbf{GGA}} & \multicolumn{1}{l|}{\textbf{SAA}} &                                          \\ \cline{1-3}
\multicolumn{1}{|l|}{\textbf{Swiss Municipalities}}               & \multicolumn{1}{l|}{840,140}      & \multicolumn{1}{l|}{210,000}      &                                          \\ \cline{1-3}
\multicolumn{1}{|l|}{\textbf{American Community Survey, 2015}}         & \multicolumn{1}{l|}{2,550,510}    & \multicolumn{1}{l|}{459,000}      &                                          \\ \cline{1-3}
\multicolumn{1}{|l|}{\textbf{US Census, 2000}}                          & \multicolumn{1}{l|}{10,872}       & \multicolumn{1}{l|}{36,000}       &                                          \\ \cline{1-3}
\multicolumn{1}{|l|}{\textbf{Kiva Loans}}                         & \multicolumn{1}{l|}{2,190,730}    & \multicolumn{1}{l|}{730,000}      &                                          \\ \cline{1-3}
\multicolumn{1}{|l|}{\textbf{UN Commodity Trade Statistics data}} & \multicolumn{1}{l|}{2,395,026}    & \multicolumn{1}{l|}{1,539,000}    &                                          \\ \cline{1-3}
                                                                  &                                   &                                   &                                          \\ \hline
\multicolumn{1}{|l|}{}                                            & \multicolumn{3}{l|}{\textbf{Average execution time per solution (seconds)}}                                     \\ \hline
\multicolumn{1}{|l|}{\textbf{Data set}}                            & \multicolumn{1}{l|}{\textbf{GGA}} & \multicolumn{1}{l|}{\textbf{SAA}} & \multicolumn{1}{l|}{\textbf{Proportion}} \\ \hline
\multicolumn{1}{|l|}{\textbf{Swiss Municipalities}}               & \multicolumn{1}{l|}{0.0009}       & \multicolumn{1}{l|}{0.0012}       & \multicolumn{1}{l|}{1.3210}              \\ \hline
\multicolumn{1}{|l|}{\textbf{American Community Survey, 2015}}         & \multicolumn{1}{l|}{0.0052}       & \multicolumn{1}{l|}{0.0011}       & \multicolumn{1}{l|}{0.2188}              \\ \hline
\multicolumn{1}{|l|}{\textbf{US Census, 2000}}                          & \multicolumn{1}{l|}{0.2177}       & \multicolumn{1}{l|}{0.0206}       & \multicolumn{1}{l|}{0.0946}              \\ \hline
\multicolumn{1}{|l|}{\textbf{Kiva Loans}}                         & \multicolumn{1}{l|}{0.0072}       & \multicolumn{1}{l|}{0.0009}       & \multicolumn{1}{l|}{0.1272}              \\ \hline
\multicolumn{1}{|l|}{\textbf{UN Commodity Trade Statistics data}} & \multicolumn{1}{l|}{0.0027}       & \multicolumn{1}{l|}{0.0008}       & \multicolumn{1}{l|}{0.2784}              \\ \hline
                                                                  &                                   &                                   &                                         
\end{tabular}
\end{table}
The above results indicate that the GGA has evaluated more solutions to find a solution of similar quality to the SAA in all cases, except for the \emph{US Census, 2000} experiment. However, we also can see that the SAA takes less time to evaluate each solution in all cases except for the \emph{Swiss Municipalities} experiment. The average execution time for each experiment can be considered in the context of the size of the data set, parallelisation, and the particular sets of hyperparameters used for the GGA and SAA. In addition to this, there is also memoisation in the evaluation algorithm for the GGA, and the gains obtained by delta evaluation by the SAA. \par Gains are more noticeable for larger data sets, because of the size of the solution and number of atomic strata in each stratum. As the strata get larger in size, the movement of $q$ atomic strata from one stratum to another (where $q$ is small) will have a smaller impact on solution quality and, therefore, the delta evaluation will be quicker.  

\section{Comparison with the continuous method in \emph{SamplingStrata}} \label{comparison_cont}

We also compared the SAA with the traditional genetic algorithm which \cite{ballin2020optimization} have applied to partition continuous strata. We used the target variables outlined in table \ref{Descriptions} above as both the continuous target and auxiliary variables (for clarity we outline them again in table \ref{xandy} below) along with the precision constraints outlined in table \ref{cvs} (the appendix). In practice, the target variable would not be exactly equal to the auxiliary variable though it is common for the auxiliary variable to be an imperfect version (for example an out-of-date or a related variable) available on the sampling frame. We invite the reader to consider this when reviewing the results of the comparisons below. It is also worth noting that initial solutions were created for both algorithms using the k-means method. Details on the training of hyperparameters for these experiments also can be found in the appendix - section \ref{Appendix}.

\begin{table}[H]
\tiny
\centering
\caption{\textbf{Summary by data set of the target and auxiliary variable descriptions for the continuous method}}
\label{xandy}
\begin{tabular}{|l|l|l|l|}
\hline
\textbf{Dataset} & \textbf{Target variables} & \textbf{Auxiliary variables} & \textbf{Description} \\ \hline
\textbf{SwissMunicipalities} & Surfacebois & Surfacebois & wood area \\ \hline
\textbf{} & Airbat & Airbat & area with buildings \\ \hline
\textbf{\begin{tabular}[c]{@{}l@{}}American Community \\ Survey, 2015\end{tabular}} & HINCP & HINCP & \begin{tabular}[c]{@{}l@{}}Household income \\ (past 12 months)\end{tabular} \\ \hline
\textbf{} & VALP & VALP & Property value \\ \hline
\textbf{} & SMOCP & SMOCP & \begin{tabular}[c]{@{}l@{}}Selected monthly \\ owner costs\end{tabular} \\ \hline
\textbf{} & INSP & INSP & \begin{tabular}[c]{@{}l@{}}Fire/hazard/flood \\ insurance (yearly amount)\end{tabular} \\ \hline
\textbf{US Census, 2000} & HHINCOME & HHINCOME & \begin{tabular}[c]{@{}l@{}}total household \\ income\end{tabular} \\ \hline
\textbf{Kiva Loans} & term\_in\_months & term\_in\_months & \begin{tabular}[c]{@{}l@{}}duration for which the \\ loan was disbursed\end{tabular} \\ \hline
\textbf{} & lender\_count & lender\_count & \begin{tabular}[c]{@{}l@{}}the total number of \\ lenders\end{tabular} \\ \hline
\textbf{} & loan & loan & the amount in USD \\ \hline
\textbf{\begin{tabular}[c]{@{}l@{}}UN Commodity Trade \\ Statistics data\end{tabular}} & trade\_usd & trade\_usd & value of the trade in USD \\ \hline
\end{tabular}
\end{table}

The attained sample sizes are compared in table \ref{continuoussamples} below where the sample size for the SAA is expressed as a ratio of the TGA. After the hyperparameters were fine-tuned (see section \ref{finetunecont}) the resulting sample sizes are comparable.    

\begin{table}[H]
\caption{\textbf{Ratio comparison of the sample sizes for the traditional genetic algorithm and simulated annealing algorithm on the continuous method}}
\label{continuoussamples}
\tiny
\centering
\begin{tabular}{|l|l|l|l|}
\hline
\textbf{Data set}                            & \textbf{TGA} & \textbf{SAA} & \textbf{Ratio} \\ \hline
\textbf{Swiss Municipalities}                & 128.69       & 120.00       & 0.93           \\ \hline
\textbf{American Community Survey, 2015}    & 4,197.68     & 3,915.48     & 0.93           \\ \hline
\textbf{US Census, 2000}                          & 192.71       & 179.89       & 0.93           \\ \hline
\textbf{Kiva Loans}                         & 3,062.33     & 3,017.79     & 0.99           \\ \hline
\textbf{UN Commodity Trade Statistics data} & 3,619.42     & 3,258.52     & 0.90           \\ \hline
\end{tabular}
\end{table}

Table \ref{continuoustimes} compares the execution times for the set of hyperparameters that found the sample sizes for each algorithm in table \ref{continuoussamples} above, as well as the total execution times taken to train that set of hyperparameters. 

\begin{table}[H]
\caption{\textbf{Ratio comparison of the execution times and total execution times for the traditional genetic algorithm and simulated annealing algorithm on the continuous method}}
\label{continuoustimes}
\tiny
\centering
\begin{tabular}{|l|l|l|l|l|l|l|}
\hline
\textbf{} &
  \multicolumn{2}{l|}{\textbf{TGA}} &
  \multicolumn{2}{l|}{\textbf{SAA}} &
  \multicolumn{2}{l|}{\textbf{Ratio comparison}} \\ \hline
\textbf{Data set} &
  \textbf{\begin{tabular}[c]{@{}l@{}l@{}l@{}}Execution \\ time\\(seconds)\end{tabular}} &
  \textbf{\begin{tabular}[c]{@{}l@{}l@{}l@{}}Total \\execution \\ time\\(seconds)\end{tabular}} &
  \textbf{\begin{tabular}[c]{@{}l@{}l@{}l@{}}Execution \\ time\\(seconds)\end{tabular}} &
  \textbf{\begin{tabular}[c]{@{}l@{}l@{}l@{}}Total \\execution \\ time\\(seconds)\end{tabular}} &
  \textbf{\begin{tabular}[c]{@{}l@{}l@{}l@{}}Execution \\ time\\(seconds)\end{tabular}} &
  \textbf{\begin{tabular}[c]{@{}l@{}l@{}l@{}}Total\\ execution \\ time\\(seconds)\end{tabular}} \\ \hline
\textbf{Swiss Municipalities}               & 753.82    & 10,434.30  & 213.44 & 1,905.82 & .28  & .18  \\ \hline
\textbf{\begin{tabular}[c]{@{}l@{}}American Community\\ Survey, 2015\end{tabular}} &
  22,016.95 &
  227,635.51 &
  13,351.19 &
  169,115.92 &
  .61 &
  .74 \\ \hline
\textbf{US Census, 2000}                          & 3,361.90  & 46,801.78  & 51.94  & 1,147.36 & .02  & .02  \\ \hline
\textbf{Kiva Loans}                         & 3,232.78  & 48,746.61  & 300.16 & 4,149.06 & .09  & .09  \\ \hline
\textbf{UN Commodity Trade Statistics data} & 29,045.23 & 326,931.63 & 73.18  & 1,287.38 & .003 & .004 \\ \hline
\end{tabular}
\end{table}
These results indicate a significantly lower execution time for the SAA for the attained solution quality. The computational efficiency gained by delta evaluation in the training of the recommended hyperparameters is also evident in the total execution times. For the \emph{American Community Survey, 2015} experiment significantly more solutions were generated by the SAA than the TGA as a result of the given hyperparameters and this impacts the execution and total execution times (see also table \ref{continuoussolutions}). Table \ref{continuoussolutions} compares the number of solutions generated by the traditional genetic algorithm with the simulated annealing algorithm. 

\begin{table}[H]
\caption{\textbf{Comparison of the number of solutions generated by the traditional genetic algorithm and simulated annealing algorithm on the continuous method}}
\label{continuoussolutions}
\tiny
\centering
\begin{tabular}{|l|l|l}
\hline
                                            & \multicolumn{2}{l|}{\textbf{Number of solutions evaluated}} \\ \hline
\textbf{Data set}                            & \textbf{TGA}       & \multicolumn{1}{l|}{\textbf{SAA}}      \\ \hline
\textbf{Swiss Municipalities}               & 840,140            & \multicolumn{1}{l|}{175,000}           \\ \hline
\textbf{American Community Survey, 2015}    & 918,102            & \multicolumn{1}{l|}{5,100,000}         \\ \hline
\textbf{US Census, 2000}                    & 43,272             & \multicolumn{1}{l|}{18,000}            \\ \hline
\textbf{Kiva Loans}                         & 146,730            & \multicolumn{1}{l|}{292,000}           \\ \hline
\textbf{UN Commodity Trade Statistics data} & 20,521,026         & \multicolumn{1}{l|}{85,500}            \\ \hline
\end{tabular}
\end{table}

In all cases except for \emph{Kiva Loans} and the \emph{American Community Survey, 2015} the SAA has generated fewer solutions. The low number of solutions generated by both algorithms for the \emph{US Census, 2000} experiment may indicate that the initial k-means solution was near the global minimum. The \emph{American Community Survey, 2015} results indicate that the SAA generated significantly more solutions to get to a comparable sample size with the TGA. As we are moving, predominantly, $q=1$ atomic strata between strata such changes in this case had limited impact on solution quality from one solution to the next. However, the gains achieved by delta evaluation meant that more solutions were evaluated per second leading to a more complete search and a lower sample size being attained.\par
For these experiments, the TGA took longer to find a comparable sample size in all cases. As pointed out in \cite{oluing2019grouping}, traditional genetic algorithms are not as efficient for grouping problems as the grouping genetic algorithm because solutions tend to have a great deal of redundancy. We would, therefore, propose that the GGA be applied also to continuous strata. On the basis of the above analysis, and the performance of SAAs in local search generally speaking along with the added gains in efficiency from delta evaluation, we would also propose that the SAA be considered as an alternative to the traditional genetic algorithm.

\section{Conclusions} \label{conclusions}

We compared the SAA with the GGA in the case of atomic strata and the TGA in the case of continuous strata \citep{ballin2020optimization}. The k-means algorithm provided good starting points in all cases. When the hyperparameters have been fine-tuned all algorithms attain results of similar quality. \par However, the execution times for the recommended hyperparameters are lower for the SAA than for the GGA with respect to atomic strata and traditional genetic algorithm with respect to continuous strata. Delta evaluation also has advantages in reducing the training times needed to find the suitable hyperparameters for the SAA. 

The GGA might benefit from being extended into a memetic algorithm by using local search to quickly improve a chromosome before adding it to the GGA chromosome population. 

The SAA, by using local search (along with a probabilistic acceptance of inferior solutions), is well suited to navigation out of local minima and the implementation of delta evaluation enables a more complete search of the local neighbourhood than would otherwise be possible in the same computation time.

\section{Further work} \label{further}

The perturbation used by the SAA randomly moves $q$ atomic strata, where mainly $q=1$, from one stratum to another. This stochastic process is standard in default simulated annealing algorithms. However, as we are using a starting solution where there is already similarity within the strata, this random process could easily move an atomic stratum ($q=1$) to a stratum where it is less suited than the stratum it was in. This suggests the presence of a certain amount of redundancy in the search for the global minimum. 

\cite{lisic2018optimal} conjecture that the introduction of nonuniform weighting in atomic strata selection could greatly improve performance of (their proposed) simulated annealing method by exchanging atomic strata near stratum boundaries more frequently than more important atomic strata. We agree that, for this algorithm, it would be more beneficial if there was a higher probability that an atomic stratum which was dissimilar to the other atomic strata was selected. We could then search for a more suitable stratum to move this atomic stratum to. 

To achieve this we could first randomly select a stratum, and then measure the Euclidean distance of each atomic stratum from that stratum medoid, weighting the chance of selection of the atomic strata in accordance with their distance from the medoid.  At this point, an atomic stratum is selected using these weighted probabilities.

The next step would be to use a K-nearest-neighbour algorithm to find the stratum medoid closest to that atomic stratum and move it to that stratum. This simple machine learning algorithm uses distance measures to classify objects based on their $K$ nearest neighbours. In this case, $k=1$, so the algorithm in practice is a closest nearest neighbour classifier.  

This additional degree of complexity to the algorithm may offset the gains achieved by using delta evaluation, particularly as the problem grows in size, thus reducing the number of solutions evaluated in the same running time. It might be more effective to use the column medians as an equivalent to the medoids. This could assist the algorithm find better quality solutions. \par However, the above suggestions may only be effective at an advanced stage of the search, where the atomic strata in each stratum are already quite similar. 

\section*{Acknowledgements}

We wish to acknowledge the editorial staff and reviewers of Survey Methodology for their constructive suggestions in the review process for this journal submission, in particular the suggestion to compare the SAA with the traditional genetic algorithm in \cite{ballin2020optimization} in the case of continuous strata. This material is based upon work supported by the Insight Centre for Data Analytics and Science Foundation Ireland under Grant No. 12/RC/2289-P2 which is co-funded under the European Regional Development Fund. Also, this publication has emanated from research supported in part by a grant from Science Foundation Ireland under Grant number 16/RC/3918 which is co-funded under the European Regional Development Fund.

%----------------------------------------------------------------------------------------
%	BIBLIOGRAPHY
%----------------------------------------------------------------------------------------

%\section*{References}
\bibliographystyle{chicago}

%\RaggedRight
\Urlmuskip=0mu plus 1mu\relax
\def\UrlBreaks{\do\/\do-}
%\bibliography{arxiv2} % The file containing the bibliography

%----------------------------------------------------------------------------------------
\newpage
\section{Appendix - Background details on the comparisons in sections \ref{experiments} and \ref{comparison_cont}}\label{Appendix}

\subsection{Precision constraints}\label{prec}

The target upper precision levels for these experiments, i.e. coefficients of variation, for each of the five experiments are provided in table \ref{cvs} below. 

\begin{table}[H]
\caption{\textbf{Summary by data set of the upper limits for the coefficients of variation}}
\label{cvs}
\tiny
\centering
\begin{tabular}{|l|l|}
\hline
   \textbf{Data set}                                         & \textbf{CV} \\ \hline
\textbf{Swiss Municipalities}                & 0.1         \\ \hline
\textbf{American Community Survey, 2015}    & 0.05        \\ \hline
\textbf{US Census, 2000}                          & 0.05        \\ \hline
\textbf{Kiva Loans}                         & 0.05        \\ \hline
\textbf{UN Commodity Trade Statistics data} & 0.05        \\ \hline
\end{tabular}
\end{table}

We selected an upper precision level of $0.1$ for the \emph{Swiss Municipalities} data set in keeping with the level set for the experiment in \cite{ballin2020optimization}. We used an upper precision level of $0.05$ for the remaining experiments, given that the upper CV levels generally set by national statistics institutes (NSIs) tend to be between $0.01$ and $0.1$, and, for this reason, results for CVs in the mid-point of this range are of interest. \par 

\subsection{Processing platform}

Table \ref{specifications} below provides details of the processing platform used for these experiments. 

\begin{table}[H]
\caption{\textbf{Specifications of the processing platform}}
\label{specifications}
\tiny
\centering
\begin{tabular}{|l|l|l|}
\hline
\textbf{Specification}                   & \textbf{Details}                              & \textbf{Notes} \\ \hline
\textbf{Processor}                       & AMD Ryzen 9 3950X 16-Core Processor, 3493 Mhz &                \\ \hline
\textbf{Cores}                           & 16 Core(s)                                    &                \\ \hline
\textbf{Logical processors}              & 32 Logical Processor(s)                       & 32 cores in R  \\ \hline
\textbf{System model}                    & X570 GAMING X                                 &                \\ \hline
\textbf{System type}                     & x64-based PC                                  &                \\ \hline
\textbf{Installed physical memory (RAM)} & 16.0 GB                                       &                \\ \hline
\textbf{Total virtual memory}            & 35.7 GB                                       &                \\ \hline
\textbf{OS name}                         & Microsoft Windows 10 Pro                      &                \\ \hline
\end{tabular}
\end{table}

In all cases, R version 4.0 or greater was used. We used the \emph{foreach} \citep{foreach} and \emph{doParallel} \citep{doParallel} packages to run the experiments in parallel. The number of cores used in the experiments was 31 (32 less 1) and this means that in the three experiments with more than 31 domains (\emph{American Community Survey 2015, Kiva Loans, UN Commodity Trade Statistics data}) the \emph{foreach} algorithm continued to loop through the available cores until a solution had been found for all domains. 

\subsection{Hyperparameters for the grouping genetic algorithm and simulated annealing algorithm}
Tables \ref{startingparametersGGA} and \ref{startingparametersSAA} below outline the number of domains in each experiment, along with number of iterations and chromosome population size for the grouping genetic algorithm and along with the number of sequences, length of sequence, and starting temperature for the simulated annealing algorithm. Section \ref{finetune} provides details on fine-tuning the hyperparameters. For more details on the hyperparameters of the GGA we refer the reader to \cite{ballin2013joint} and \cite{oluing2019grouping} and of the SAA to sections \ref{SA} and \ref{outline}.

\begin{table}[H]
\caption{\textbf{Summary by data set of the hyperparameters for the grouping genetic algorithm for each domain}}
\label{startingparametersGGA}
\tiny
\centering
\begin{tabular}{|l|l|l|l|l|l|l|}
\hline
\textbf{Data set} &
  \textbf{Domains} &
  \textbf{\begin{tabular}[c]{@{}l@{}}Number of \\ iterations, I\end{tabular}} &
  \textbf{\begin{tabular}[c]{@{}l@{}l@{}}Chromosome\\ population \\ size, $N_p$\end{tabular}} &
  \textbf{\begin{tabular}[c]{@{}l@{}}Mutation \\ chance \end{tabular}} &
  \textbf{\begin{tabular}[c]{@{}l@{}}Elitism \\ rate, $E_R$\end{tabular}} &
  \textbf{\begin{tabular}[c]{@{}l@{}}Add strata \\ factor\end{tabular}} \\ \hline
\textbf{Swiss Municipalities}                & 7   & 4,000 & 50 & 0.0053360 & 0.4 & 0.0037620 \\ \hline
\textbf{American Community Survey, 2015}    & 51  & 5,000  & 20 & 0.0008134 & 0.5 & 0.0610529 \\ \hline
\textbf{US Census, 2000}                          & 9   & 100   & 20 & 0.0000007 & 0.4 & 0.0000472 \\ \hline
\textbf{Kiva Loans}                         & 73  & 3,000 & 20 & 0.0007221 & 0.5 & 0.0685005 \\ \hline
\textbf{UN Commodity Trade Statistics data} & 171 & 1,000 & 20 & 0.0004493 & 0.3 & 0.0866266 \\ \hline
\end{tabular}
\end{table}

\begin{table}[H]
\caption{\textbf{Summary by data set of the hyperparameters for the simulated annealing algorithm for each domain}}
\label{startingparametersSAA}
\tiny
\centering
\begin{tabular}{|l|l|l|l|l|l|l|l|}
\hline
\textbf{Data set} &
  \textbf{Domains} &
  \textbf{\begin{tabular}[c]{@{}l@{}}Number of\\ sequences,\\ $maxit$\end{tabular}} &
  \textbf{\begin{tabular}[c]{@{}l@{}}Length of\\ sequence,\\ $J$ \end{tabular}} &
  \textbf{Temperature, $T$} &
  \textbf{\begin{tabular}[c]{@{}l@{}}Decrement\\ constant, $DC$\end{tabular}} &
  \textbf{\begin{tabular}[c]{@{}l@{}}\% of L for \\ maximum \\ q value,$L_{max \%}$\end{tabular}} &
  \textbf{\begin{tabular}[c]{@{}l@{}l@{}}Probability\\ of new\\ stratum,\\ $P(H+1)$ \end{tabular}} \\ \hline
\textbf{Swiss Municipalities}               & 7   & 10 & 3,000 & 0.0000720 & 0.5083686 & 0.0183356 & 0.0997907 \\ \hline
\textbf{\begin{tabular}[c]{@{}l@{}}American Community \\ Survey, 2015\end{tabular}}     & 51  & 3  & 3,000 & 0.0002347 & 0.6873029 & 0.0076477 & 0.0291729 \\ \hline
\textbf{US Census, 2000}                          & 9   & 2  & 2,000 & 0.0006706 & 0.5457192 & 0.0189395 & 0.0806919 \\ \hline
\textbf{Kiva Loans}                         & 73  & 5  & 2,000 & 0.0009935 & 0.7806557 & 0.0143925 & 0.0317491 \\ \hline
\textbf{\begin{tabular}[c]{@{}l@{}}UN Commodity Trade \\ Statistics data\end{tabular}} & 171 & 3  & 3,000 & 0.0007902 & 0.5072737 & 0.0234728 & 0.0013775 \\ \hline
\end{tabular}
\end{table}

\subsection{Fine-tuning the hyperparameters for the grouping genetic algorithm and simulated annealing algorithm}
\label{finetune}
In order to fine-tune the initial parameters or \emph{hyperparameters} we used sequential model-based optimization \citep{hutter2010sequential}. We first generated an initial design of hyperparameters from the value ranges described for the GGA in table \ref{GGAvalues} and in table \ref{SAAvalues} for the SAA below using the latin hypercube design method \citep{mckay2000comparison}.

\begin{table}[H]
\caption{\textbf{Ranges for fine-tuning the hyperparameters for the grouping genetic algorithm}}
\label{GGAvalues}
\tiny
\centering
\begin{tabular}{llllllll}
\cline{1-7}
\multicolumn{1}{|l|}{} &
  \multicolumn{3}{l|}{\textbf{Iterations}} &
  \multicolumn{3}{l|}{\textbf{Population size}} &
   \\ \cline{1-7}
\multicolumn{1}{|l|}{\textbf{Value type}} &
  \multicolumn{3}{l|}{\textbf{Discrete}} &
  \multicolumn{3}{l|}{\textbf{Discrete}} &
   \\ \cline{1-7}
\multicolumn{1}{|l|}{\textbf{Value range}} &
  \multicolumn{1}{l|}{\textbf{\begin{tabular}[c]{@{}l@{}}Lower \\ value\end{tabular}}} &
  \multicolumn{1}{l|}{\textbf{\begin{tabular}[c]{@{}l@{}}Upper \\ value\end{tabular}}} &
  \multicolumn{1}{l|}{\textbf{Increments}} &
  \multicolumn{1}{l|}{\textbf{\begin{tabular}[c]{@{}l@{}}Lower \\ value\end{tabular}}} &
  \multicolumn{1}{l|}{\textbf{\begin{tabular}[c]{@{}l@{}}Upper \\ value\end{tabular}}} &
  \multicolumn{1}{l|}{\textbf{Increments}} &
   \\ \cline{1-7}
\multicolumn{1}{|l|}{\textbf{Swiss Municipalities}} &
  \multicolumn{1}{l|}{500} &
  \multicolumn{1}{l|}{5,000} &
  \multicolumn{1}{l|}{500} &
  \multicolumn{1}{l|}{10} &
  \multicolumn{1}{l|}{50} &
  \multicolumn{1}{l|}{10} &
   \\ \cline{1-7}
\multicolumn{1}{|l|}{\textbf{\begin{tabular}[c]{@{}l@{}}American Community \\ Survey, 2015\end{tabular}}} &
  \multicolumn{1}{l|}{1,000} &
  \multicolumn{1}{l|}{5,000} &
  \multicolumn{1}{l|}{1,000} &
  \multicolumn{1}{l|}{10} &
  \multicolumn{1}{l|}{20} &
  \multicolumn{1}{l|}{10} &
   \\ \cline{1-7}
\multicolumn{1}{|l|}{\textbf{Kiva Loans}} &
  \multicolumn{1}{l|}{1,000} &
  \multicolumn{1}{l|}{3,000} &
  \multicolumn{1}{l|}{1,000} &
  \multicolumn{1}{l|}{10} &
  \multicolumn{1}{l|}{20} &
  \multicolumn{1}{l|}{10} &
   \\ \cline{1-7}
\multicolumn{1}{|l|}{\textbf{\begin{tabular}[c]{@{}l@{}}UN Commodity Trade \\ Statistics data\end{tabular}}} &
  \multicolumn{1}{l|}{500} &
  \multicolumn{1}{l|}{1,000} &
  \multicolumn{1}{l|}{500} &
  \multicolumn{1}{l|}{10} &
  \multicolumn{1}{l|}{20} &
  \multicolumn{1}{l|}{10} &
   \\ \cline{1-7}
\multicolumn{1}{|l|}{\textbf{US Census, 2000}} &
  \multicolumn{1}{l|}{50} &
  \multicolumn{1}{l|}{100} &
  \multicolumn{1}{l|}{50} &
  \multicolumn{1}{l|}{10} &
  \multicolumn{1}{l|}{20} &
  \multicolumn{1}{l|}{10} &
   \\ \cline{1-7}
 &
   &
   &
   &
   &
   &
   &
   \\ \hline
\multicolumn{1}{|l|}{} &
  \multicolumn{2}{l|}{\textbf{Mutation chance}} &
  \multicolumn{3}{l|}{\textbf{Elitism rate, $E_R$}} &
  \multicolumn{2}{l|}{\textbf{Add strata factor}} \\ \hline
\multicolumn{1}{|l|}{\textbf{Value type}} &
  \multicolumn{2}{l|}{\textbf{Numeric}} &
  \multicolumn{3}{l|}{\textbf{Discrete}} &
  \multicolumn{2}{l|}{\textbf{Numeric}} \\ \hline
\multicolumn{1}{|l|}{\textbf{Value range}} &
  \multicolumn{1}{l|}{\textbf{\begin{tabular}[c]{@{}l@{}}Lower \\ value\end{tabular}}} &
  \multicolumn{1}{l|}{\textbf{\begin{tabular}[c]{@{}l@{}}Upper \\ value\end{tabular}}} &
  \multicolumn{1}{l|}{\textbf{\begin{tabular}[c]{@{}l@{}}Lower \\ value\end{tabular}}} &
  \multicolumn{1}{l|}{\textbf{\begin{tabular}[c]{@{}l@{}}Upper \\ value\end{tabular}}} &
  \multicolumn{1}{l|}{\textbf{Increments}} &
  \multicolumn{1}{l|}{\textbf{\begin{tabular}[c]{@{}l@{}}Lower \\ value\end{tabular}}} &
  \multicolumn{1}{l|}{\textbf{\begin{tabular}[c]{@{}l@{}}Upper \\ value\end{tabular}}} \\ \hline
\multicolumn{1}{|l|}{\textbf{Swiss Municipalities}} &
  \multicolumn{1}{l|}{0} &
  \multicolumn{1}{l|}{0.10} &
  \multicolumn{1}{l|}{0.1} &
  \multicolumn{1}{l|}{0.5} &
  \multicolumn{1}{l|}{0.1} &
  \multicolumn{1}{l|}{0} &
  \multicolumn{1}{l|}{0.1} \\ \hline
\multicolumn{1}{|l|}{\textbf{American Community Survey, 2015}} &
  \multicolumn{1}{l|}{0} &
  \multicolumn{1}{l|}{0.001} &
  \multicolumn{1}{l|}{0.1} &
  \multicolumn{1}{l|}{0.5} &
  \multicolumn{1}{l|}{0.1} &
  \multicolumn{1}{l|}{0} &
  \multicolumn{1}{l|}{0.1} \\ \hline
\multicolumn{1}{|l|}{\textbf{Kiva Loans}} &
  \multicolumn{1}{l|}{0} &
  \multicolumn{1}{l|}{0.001} &
  \multicolumn{1}{l|}{0.1} &
  \multicolumn{1}{l|}{0.5} &
  \multicolumn{1}{l|}{0.1} &
  \multicolumn{1}{l|}{0} &
  \multicolumn{1}{l|}{0.1} \\ \hline
\multicolumn{1}{|l|}{\textbf{UN Commodity Trade Statistics data}} &
  \multicolumn{1}{l|}{0} &
  \multicolumn{1}{l|}{0.001} &
  \multicolumn{1}{l|}{0.1} &
  \multicolumn{1}{l|}{0.5} &
  \multicolumn{1}{l|}{0.1} &
  \multicolumn{1}{l|}{0} &
  \multicolumn{1}{l|}{0.1} \\ \hline
\multicolumn{1}{|l|}{\textbf{US Census, 2000}} &
  \multicolumn{1}{l|}{0} &
  \multicolumn{1}{l|}{0.000001} &
  \multicolumn{1}{l|}{0.1} &
  \multicolumn{1}{l|}{0.5} &
  \multicolumn{1}{l|}{0.1} &
  \multicolumn{1}{l|}{0} &
  \multicolumn{1}{l|}{0.0001} \\ \hline
\end{tabular}
\end{table}

\begin{table}[H]
\caption{\textbf{Ranges for fine-tuning the hyperparameters for the simulated annealing algorithm}}
\label{SAAvalues}
\tiny
\centering
\begin{tabular}{lllllllll}
\cline{1-7}
\multicolumn{1}{|l|}{} &
  \multicolumn{3}{l|}{\textbf{\begin{tabular}[c]{@{}l@{}}Number of \\ sequences, maxit\end{tabular}}} &
  \multicolumn{3}{l|}{\textbf{\begin{tabular}[c]{@{}l@{}}Length of \\ sequence, J\end{tabular}}} &
   &
   \\ \cline{1-7}
\multicolumn{1}{|l|}{\textbf{Value type}} &
  \multicolumn{3}{l|}{\textbf{Discrete}} &
  \multicolumn{3}{l|}{\textbf{Discrete}} &
   &
   \\ \cline{1-7}
\multicolumn{1}{|l|}{\textbf{Value range}} &
  \multicolumn{1}{l|}{\textbf{\begin{tabular}[c]{@{}l@{}}Lower \\ value\end{tabular}}} &
  \multicolumn{1}{l|}{\textbf{\begin{tabular}[c]{@{}l@{}}Upper \\ value\end{tabular}}} &
  \multicolumn{1}{l|}{\textbf{Increments}} &
  \multicolumn{1}{l|}{\textbf{\begin{tabular}[c]{@{}l@{}}Lower \\ value\end{tabular}}} &
  \multicolumn{1}{l|}{\textbf{\begin{tabular}[c]{@{}l@{}}Upper \\ value\end{tabular}}} &
  \multicolumn{1}{l|}{\textbf{Increments}} &
   &
   \\ \cline{1-7}
\multicolumn{1}{|l|}{\textbf{Swiss Municipalities}} &
  \multicolumn{1}{l|}{10} &
  \multicolumn{1}{l|}{50} &
  \multicolumn{1}{l|}{10} &
  \multicolumn{1}{l|}{1,000} &
  \multicolumn{1}{l|}{3,000} &
  \multicolumn{1}{l|}{1,000} &
   &
   \\ \cline{1-7}
\multicolumn{1}{|l|}{\textbf{\begin{tabular}[c]{@{}l@{}}American Community \\ Survey, 2015\end{tabular}}} &
  \multicolumn{1}{l|}{1} &
  \multicolumn{1}{l|}{3} &
  \multicolumn{1}{l|}{1} &
  \multicolumn{1}{l|}{1,000} &
  \multicolumn{1}{l|}{3,000} &
  \multicolumn{1}{l|}{1,000} &
   &
   \\ \cline{1-7}
\multicolumn{1}{|l|}{\textbf{Kiva Loans}} &
  \multicolumn{1}{l|}{1} &
  \multicolumn{1}{l|}{5} &
  \multicolumn{1}{l|}{1} &
  \multicolumn{1}{l|}{1,000} &
  \multicolumn{1}{l|}{2,000} &
  \multicolumn{1}{l|}{1,000} &
   &
   \\ \cline{1-7}
\multicolumn{1}{|l|}{\textbf{\begin{tabular}[c]{@{}l@{}}UN Commodity Trade \\ Statistics data\end{tabular}}} &
  \multicolumn{1}{l|}{1} &
  \multicolumn{1}{l|}{3} &
  \multicolumn{1}{l|}{1} &
  \multicolumn{1}{l|}{1,000} &
  \multicolumn{1}{l|}{3,000} &
  \multicolumn{1}{l|}{1,000} &
   &
   \\ \cline{1-7}
\multicolumn{1}{|l|}{\textbf{US Census, 2000}} &
  \multicolumn{1}{l|}{1} &
  \multicolumn{1}{l|}{2} &
  \multicolumn{1}{l|}{1} &
  \multicolumn{1}{l|}{1,000} &
  \multicolumn{1}{l|}{2,000} &
  \multicolumn{1}{l|}{1,000} &
   &
   \\ \cline{1-7}
 &
   &
   &
   &
   &
   &
   &
   &
   \\ \hline
\multicolumn{1}{|l|}{} &
  \multicolumn{2}{l|}{\textbf{Temperature, T}} &
  \multicolumn{2}{l|}{\textbf{\begin{tabular}[c]{@{}l@{}}Decrement \\ constant, DC\end{tabular}}} &
  \multicolumn{2}{l|}{\textbf{\begin{tabular}[c]{@{}l@{}}\% L for\\ maximum q\\ value, $L_{max \%}$\end{tabular}}} &
  \multicolumn{2}{l|}{\textbf{\begin{tabular}[c]{@{}l@{}}Probability \\ of \\ new stratum, P(H+1)\end{tabular}}} \\ \hline
\multicolumn{1}{|l|}{\textbf{Value type}} &
  \multicolumn{2}{l|}{\textbf{Numeric}} &
  \multicolumn{2}{l|}{\textbf{Numeric}} &
  \multicolumn{2}{l|}{\textbf{Numeric}} &
  \multicolumn{2}{l|}{\textbf{Numeric}} \\ \hline
\multicolumn{1}{|l|}{\textbf{Value Range}} &
  \multicolumn{1}{l|}{\textbf{\begin{tabular}[c]{@{}l@{}}Lower \\ value\end{tabular}}} &
  \multicolumn{1}{l|}{\textbf{\begin{tabular}[c]{@{}l@{}}Upper \\ value\end{tabular}}} &
  \multicolumn{1}{l|}{\textbf{\begin{tabular}[c]{@{}l@{}}Lower \\ value\end{tabular}}} &
  \multicolumn{1}{l|}{\textbf{\begin{tabular}[c]{@{}l@{}}Upper \\ value\end{tabular}}} &
  \multicolumn{1}{l|}{\textbf{\begin{tabular}[c]{@{}l@{}}Lower \\ value\end{tabular}}} &
  \multicolumn{1}{l|}{\textbf{\begin{tabular}[c]{@{}l@{}}Upper \\ value\end{tabular}}} &
  \multicolumn{1}{l|}{\textbf{\begin{tabular}[c]{@{}l@{}}Lower \\ value\end{tabular}}} &
  \multicolumn{1}{l|}{\textbf{\begin{tabular}[c]{@{}l@{}}Upper \\ value\end{tabular}}} \\ \hline
\multicolumn{1}{|l|}{\textbf{Swiss Municipalities}} &
  \multicolumn{1}{l|}{0} &
  \multicolumn{1}{l|}{0.001} &
  \multicolumn{1}{l|}{0.5} &
  \multicolumn{1}{l|}{1} &
  \multicolumn{1}{l|}{0.0001} &
  \multicolumn{1}{l|}{0.025} &
  \multicolumn{1}{l|}{0} &
  \multicolumn{1}{l|}{0.1} \\ \hline
\multicolumn{1}{|l|}{\textbf{\begin{tabular}[c]{@{}l@{}}American Community \\ Survey, 2015\end{tabular}}} &
  \multicolumn{1}{l|}{0} &
  \multicolumn{1}{l|}{0.001} &
  \multicolumn{1}{l|}{0.5} &
  \multicolumn{1}{l|}{1} &
  \multicolumn{1}{l|}{0.0001} &
  \multicolumn{1}{l|}{0.025} &
  \multicolumn{1}{l|}{0} &
  \multicolumn{1}{l|}{0.1} \\ \hline
\multicolumn{1}{|l|}{\textbf{Kiva Loans}} &
  \multicolumn{1}{l|}{0} &
  \multicolumn{1}{l|}{0.001} &
  \multicolumn{1}{l|}{0.5} &
  \multicolumn{1}{l|}{1} &
  \multicolumn{1}{l|}{0.0001} &
  \multicolumn{1}{l|}{0.025} &
  \multicolumn{1}{l|}{0} &
  \multicolumn{1}{l|}{0.1} \\ \hline
\multicolumn{1}{|l|}{\textbf{\begin{tabular}[c]{@{}l@{}}UN Commodity Trade \\ Statistics data\end{tabular}}} &
  \multicolumn{1}{l|}{0} &
  \multicolumn{1}{l|}{0.001} &
  \multicolumn{1}{l|}{0.5} &
  \multicolumn{1}{l|}{1} &
  \multicolumn{1}{l|}{0.0001} &
  \multicolumn{1}{l|}{0.025} &
  \multicolumn{1}{l|}{0} &
  \multicolumn{1}{l|}{0.1} \\ \hline
\multicolumn{1}{|l|}{\textbf{US Census, 2000}} &
  \multicolumn{1}{l|}{0} &
  \multicolumn{1}{l|}{0.001} &
  \multicolumn{1}{l|}{0.5} &
  \multicolumn{1}{l|}{1} &
  \multicolumn{1}{l|}{0.0001} &
  \multicolumn{1}{l|}{0.025} &
  \multicolumn{1}{l|}{0} &
  \multicolumn{1}{l|}{0.1} \\ \hline
\textbf{} &
   &
   &
   &
   &
   &
   &
   &
\end{tabular}
\end{table}

As some of the hyperparameter value ranges were discrete, we used a random forest with regression trees to develop a surrogate learner model. After this, a confidence bound using a lambda value, $\lambda$, to control the trade-off between exploitation and exploration was used as the acquisition function. The focus search approach \citep{bischla2017mlrmbo} was used to optimise the acquisition function which, in turn, was used to propose the hyperparameters which were evaluated using the surrogate function (which is a cheaper alternative to using the GGA or SAA algorithms). From these, the most promising hyperparameters were then evaluated by the GGA or SAA and the hyperparameters and solution costs added to the initial design. The process was then repeated for a set number of iterations and the best performing hyperparameters and solution outcomes were selected. We implemented this using the \emph{MBO} function with the parameters outlined in table \ref{MBOparameters}. These are distinct from the parameters being fine-tuned, which are outlined in tables \ref{GGAvalues} and \ref{SAAvalues} above. 

\begin{table}[H]
\caption{\textbf{Parameters used in the \emph{MBO} Function}}
\label{MBOparameters}
\tiny
\centering
\begin{tabular}{|l|l|}
\hline
\textbf{MBO parameters}                                      & \multicolumn{1}{c|}{\textbf{Value}} \\ \hline
\textbf{Initial Design size (Latin Hypercube Design method)} & $10$                                  \\ \hline
\textbf{Iterations, number of }                                       & $10$                                  \\ \hline
\textbf{Number of Trees}                                     & $500$                                 \\ \hline
\textbf{Lambda}, $\lambda$                                               & $5$                                   \\ \hline
\textbf{Focus Search Points}                                 & $1,000$                                \\ \hline
\end{tabular}
\end{table}

 As can be seen from the limited scope of the \emph{MBO} function parameters this was not an exhaustive fine-tuning of the hyperparameters for the GGA and SAA. The aim of these experiments was to consider whether the SAA can attain comparable solution quality with the GGA in less computation time per solution thus resulting in savings in execution times. However, we also compared the total execution times as this is a consequence of the need to train the hyperparameters for both algorithms. \par

Tables outlining the hyperparameters, in each of the 20 fine-tuning iterations, for each experiment are available from the authors on request. The first 10 sets of hyperparameters were randomly generated from the ranges laid out in tables \ref{GGAvalues} and \ref{SAAvalues}. The ranges selected were identified using practical knowledge of the algorithms and data. The second 10 sets reflects the \emph{MBO} function's attempts to learn the hyperparameters that best lead each algorithm towards the optimal solution using the previous solutions as a guide. 

\subsection{Hyperparameters for the traditional genetic algorithm and simulated annealing algorithm}

Tables \ref{continuousga} and \ref{continuousSAA} outline the hyperparameters for the tradtional genetic algorithm and the simulated annealing algorithm. The add strata factor option is not available for the traditional genetic algorithm and, therefore, is not included in table \ref{continuousga}. More details on fine-tuning the hyperparameters are provided in section \ref{finetunecont}.

\begin{table}[H]
\tiny
\centering
\caption{\textbf{Hyperparameters for the traditional genetic algorithm}}
\label{continuousga}
\begin{tabular}{|l|l|l|l|l|l|}
\hline
\textbf{Data set} & \textbf{Iterations} & \textbf{Population size} & \textbf{Mutation chance} & \textbf{Elitism rate, $E_R$} \\ \hline
\textbf{Swiss Municipalities}                & 4,000 & 50 & 0.0053360 & 0.4 \\ \hline
\textbf{American Community Survey, 2015}    & 1,000 & 20 & 0.0009952 & 0.1 \\ \hline
\textbf{US Census, 2000}                          & 400   & 20 & 0.0002317 & 0.4 \\ \hline
\textbf{Kiva Loans}                         & 200   & 20 & 0.0817285 & 0.5 \\ \hline
\textbf{UN Commodity Trade Statistics data} & 5,000 & 30 & 0.0005599 & 0.2 \\ \hline
\end{tabular}
\end{table}

\begin{table}[H]
\tiny
\centering
\caption{\textbf{Hyperparameters for the simulated annealing algorithm}}
\label{continuousSAA}
\begin{tabular}{|l|l|l|l|l|l|l|l|}
\hline
\textbf{Data set} &
  \textbf{\begin{tabular}[c]{@{}l@{}l@{}l@{}}Number of\\ sequences, \\maxit\end{tabular}} &
  \textbf{\begin{tabular}[c]{@{}l@{}l@{}l@{}}Length of\\ sequence, J\end{tabular}} &
  \textbf{Temperature, T} &
  \textbf{\begin{tabular}[c]{@{}l@{}l@{}l@{}}Decrement\\ constant , DC\end{tabular}} &
  \textbf{\begin{tabular}[c]{@{}l@{}l@{}l@{}}\% for \\maximum\\ q value,\\ $L_{max \%}$\end{tabular}} &
  \textbf{\begin{tabular}[c]{@{}l@{}l@{}l@{}}Probability\\ of new\\ stratum, \\P(H+1) \end{tabular}} \\ \hline
\textbf{Swiss Municipalities}                & 5  & 5,000 & 0.02311057 & 0.9427609 & 0.3736443 & 0.0229361 \\ \hline
\textbf{American Community Survey, 2015}    & 50 & 2,000 & 0.00000005 & 0.9528952 & 0.0001021 & 0.0000008 \\ \hline
\textbf{US Census, 2000}                          & 1  & 2,000 & 0.00002000 & 0.9665631 & 0.0221147 & 0.0160408 \\ \hline
\textbf{Kiva Loans}                         & 2  & 2,000 & 0.00053839 & 0.8660943 & 0.0014281 & 0.0216320 \\ \hline
\textbf{UN Commodity Trade Statistics data} & 2  & 250   & 0.00067481 & 0.9309940 & 0.0203113 & 0.0149499 \\ \hline
\end{tabular}
\end{table}

\subsection{Fine-tuning the hyperparameters for the traditional genetic algorithm and simulated annealing algorithm}
\label{finetunecont}

We fine-tuned the hyperparameters for the TGA and SAA using the same methodology described in section \ref{finetune}. Tables outlining the hyperparameters, in each of the 20 fine-tuning iterations, for each experiment are available from the authors on request. The first 10 sets were randomly generated using practical knowledge of the algorithms and data to define upper and lower bounds for each hyperparameter. In the second 10 sets the \emph{MBO} function attempts to optimise the hyperparameters using the previous solutions as a guide. 

\end{document}